%% file: tran.tex
\documentclass[english]{article}
\usepackage[latin9]{inputenc}
\usepackage{geometry}
\usepackage{array}
\usepackage{float}
\usepackage{amssymb}
\usepackage{graphicx}
\usepackage{epstopdf}

\usepackage{units}
\usepackage{amsmath}
\usepackage{amssymb}
\usepackage{graphicx}

\date{}

\title{Mixed-Variate Restricted Boltzmann Machines\thanks{Work done when authors were with Curtin University, Australia.}}

\author{Truyen Tran$^{\dagger\ddagger}$, Dinh Phung$^{\dagger}$, Svetha
Venkatesh$^{\dagger}$ \\
 $^{\dagger}$Pattern Recognition and Data Analytics, Deakin University,
Australia\\
 $^{\ddagger}$Department of Computing, Curtin University, Australia\\
 \{truyen.tran,dinh.phung,svetha.venkatesh\}@deakin.edu.au }

\begin{document}

\maketitle
\global\long\def\hb{\boldsymbol{h}}

\global\long\def\h{h}

\global\long\def\vb{\boldsymbol{v}}

\global\long\def\v{v}

\global\long\def\MODEL{\mathrm{MV.RBM}}

\begin{abstract}
Modern datasets are becoming heterogeneous. To this end, we present
in this paper \emph{Mixed-Variate Restricted Boltzmann Machines} for
simultaneously modelling variables of multiple types and modalities,
including \emph{binary }and\emph{ continuous }responses, \emph{categorical
}options, \emph{multicategorical }choices, \emph{ordinal }assessment
and \emph{category-ranked }preferences. Dependency among variables
is modeled using latent binary variables, each of which can be interpreted
as a particular hidden aspect of the data. The proposed model, similar
to the standard RBMs, allows fast evaluation of the posterior for
the latent variables. Hence, it is naturally suitable for many common
tasks including, but not limited to, (a) as a pre-processing step
to convert complex input data into a more convenient vectorial representation
through the latent posteriors, thereby offering a dimensionality reduction
capacity, (b) as a classifier supporting binary, multiclass, multilabel,
and label-ranking outputs, or a regression tool for continuous outputs
and (c) as a data completion tool for multimodal and heterogeneous
data. We evaluate the proposed model on a large-scale dataset using
the world opinion survey results on three tasks: feature extraction
and visualization, data completion and prediction.

\end{abstract}

\section{Introduction}

\input{intro.tex}

\section{Mixed-Variate Restricted Boltzmann Machines \label{sec:Mixed-State-Restricted}}

\input{model.tex}

\section{Learning and Inference \label{sec:Learning-and-Inference}}

\input{learning.tex}

\section{A Case Study: World Attitudes \label{sub:Filling-Missing-Data}}

\input{experiments.tex}

\section{Related Work}

The most popular use of RBMs is in modelling of individual types,
for example, binary variables \cite{freund1994unsupervised}, Gaussian
variables \cite{hinton2006rdd,ranzato2010modeling}, categorical
variables \cite{Salakhutdinov-et-alICML07}, rectifier linear units
\cite{nair2010rectified}, Poisson variables \cite{gehler2006rap},
counts \cite{salakhutdinov2009replicated} and Beta variables \cite{le2011learning}. When RBMs are used
for classification \cite{larochelle2008classification}, categorical
variables might be employed for labeling in additional to the features.
Other than that, there has been a model called Dual-Wing RBM for modelling
both continuous and binary variables \cite{xing2005mining}. However,
there have been no attempts to address all \emph{six} data types in
a single model as we do in the present paper.

The literature on ordinal variables is sufficiently rich in
statistics, especially after the seminal work of \cite{mccullagh1980rmo}.
In machine learning, on the other hand, the literature is quite sparse
and recent (e.g. see \cite{shashua2002tlm,yu2006collaborative}) and it is often limitted to single
ordinal output (given numerical input co-variates). An RBM-based modelling of ordinal variables
addressed in \cite{Truyen:2009a} is similar to ours, except that
our treatment is more general and principled.

Mixed-variate modelling has been previously studied in statistics, under
a variety of names such as \emph{mixed outcomes}, \emph{mixed data},
or \emph{mixed responses} \cite{sammel1997latent,dunson2000bayesian,shi2000latent,mcculloch2008joint}. 
Most papers focus on the mix of ordinal, Gaussian and binary variables under the
\emph{latent variable} framework. More specifically, each observed variable is
assumed to be generated from one or more underlying continuous latent variables. 
Inference becomes complicated since we need to integrate out
these correlated latent variables, making it difficult to handle hundreds of variables
and large-scale datasets.

In machine learning, the problem of predicting a single multicategorical
variable is also known as multilabel learning (e.g., see \cite{tsoumakas2007multi}).
Previous ideas that we have adapted into our context including the
shared structure among labels \cite{ji2008extracting}. In our model,
the sharing is captured by the hidden layer in a probabilistic manner
and we consider many multicategorical variables at the same time.
Finally, the problem of predicting a single category-ranked variable
is also known as label-ranking (e.g., see \cite{dekel2003llm,hullermeier2008lrl}).
The idea we adopt is the pairwise comparison between categories. However,
the previous work neither considered the hidden correlation between
those pairs nor attempted multiple category-ranked variables.

\section{Conclusion}

We have introduced Mixed-Variate Restricted Boltzmann Machines ($\MODEL$)
as a generalisation of the RBMs for modelling correlated variables
of multiple modalities and types. Six types considered were: binary,
categorical, multicategorical, continuous information, ordinal, and
category-ranking. We shown that the $\MODEL$ is capable of handling
a variety of machine learning tasks including feature exaction, dimensionality
reduction, data completion, and label prediction. We demonstrated
the capacity of the model on a large-scale world-wide survey.

We plan to further the present work in several directions. First,
the model has the capacity to handle multiple related predictive models
simultaneously by learning a shared representation through hidden
posteriors, thereby applicable to the setting of multitask learning.
Second, there may exist strong interactions between variables which
the RBM architecture may not be able to capture. The theoretical question
is then how to model inter-type dependencies directly without going
through an intermediate hidden layer. Finally, we plan to enrich the
range of applications of the proposed model.

\paragraph{Acknowledgment:}
We thank anonymous reviewers for insightful comments.

\appendix

\section{Additional Materials}

\input{appendix.tex}

\end{document}

%% file: intro.tex
Restricted Boltzmann Machines (RBM) \cite{hinton1986learning,freund1994unsupervised}
have recently attracted an increasing attention for their rich capacity
in a variety of learning tasks, including multivariate distribution
modelling, feature extraction, classification, and construction of
deep architectures \cite{hinton2006rdd,salakhutdinov2009deep}. An
RBM is a two-layer Markov random field in which the visible layer
represents observed variables and the hidden layer represents latent
aspects of the data. Pairwise interactions are only permitted for units
between layers. As a result, the posterior distribution over the
hidden variables and the probability of the data generative model
are easy to evaluate, allowing fast feature extraction and efficient
sampling-based inference \cite{Hinton02}. Nonetheless, most existing
work in RBMs implicitly assumes that the visible layer contains variables
of the same modality. By far the most popular input types are binary
\cite{freund1994unsupervised} and Gaussian \cite{hinton2006rdd}.
Recent extension includes categorical \cite{Salakhutdinov-et-alICML07},
ordinal \cite{Truyen:2009a}, Poisson \cite{gehler2006rap}
and Beta \cite{le2011learning} data.
To the best of our knowledge, \emph{none} has been considered for
multicategorical and category-ranking data, \emph{nor} for a mixed
combination of these data types.

In this paper, we investigate a generalisation of the RBM for variables
of multiple modalities and types. Take, for example, data from a typical
survey, where a person is asked a variety of questions in
many styles ranging from \textsf{yes}/\textsf{no} to multiple choices
and preference statements. Typically, there are six question/answer
types: (1) binary responses (e.g., \textsf{satisfied}\emph{ }vs.\emph{
}\textsf{unsatisfied}), (2) categorical options (e.g., one of \textsf{employed}\emph{,
}\textsf{unemployed}\emph{ }or\emph{ }\textsf{retired}), (iii) multicategorical
choices (e.g., any of \textsf{family}\emph{, }\textsf{education}\emph{
}or\emph{ }\textsf{income}), (iv) continuous information (e.g. \textsf{age}),
(v) ordinal assessment (e.g., one of \textsf{good}\emph{, }\textsf{neural}\emph{
}or\emph{ }\textsf{bad}), and (vi) category-ranked preferences (e.g.,
in the decreasing order of importance: \textsf{children}\emph{, }\textsf{security}\emph{,
}\textsf{food}\emph{ }and\emph{ }\textsf{money}). As the answers in
a response come from the same person, they are inherently correlated.
For instance, a young American is likely to own a computer, whilst
a typical Chinese adult may concern more about their children's education.
However, modelling the direct correlation among multiple types is
difficult. We show, on the other hand, a two-layer RBM is well-suited
for this problem. First, its undirected graphical structure allows
a great flexibility to encode all six data types into the same probability
distribution. Second, the binary hidden layer pools information from
visible units and redistributes to all others, thereby introducing
dependencies among variables. We term our model the \emph{Mixed-Variate
Restricted Boltzmann Machines} ($\MODEL$).

The $\MODEL$ has the capacity of supporting a variety of machine
learning tasks. Its posteriors can be used as a vectorial representation
of the data hiding away the obscured nature of the observed data.
As the result, we can use $\MODEL$ for data pre-processing, visualisation,
and dimensionality reduction. Given the hidden layer, the original
and missing observables can also be reconstructed through the
generative data model. By splitting the observed data into an input
and output sets, predictive models can be learnt to perform classification,
ranking or regression. These capacities are demonstrated in this paper
on a large-scale international opinion survey across $44$ nations
involving more than $38$ thousand people.

%% file: model.tex
In this section we present Mixed-Variate Restricted Boltzmann Machines
($\MODEL$) for jointly modelling variables of multiple modalities and
types. For ease of following the text, we include a notation description
in Table~\ref{tab:Notations}.

\begin{table}
\begin{centering}
\begin{tabular}{lllll}
\hline 
$\v_{i}$ & Single visible variable & ~~~ & $G_{i}(v_{i})$,$H_{ik}(\v_{i})$ & Functions of an input variable\tabularnewline
$\vb$ & A set of visible variables &  & $U_{i}$,$U_{im}$,$U_{id}$ & Input bias parameters\tabularnewline
$\h_{k}$ & Single hidden variable &  & $V_{ik}$,$V_{imk}$,$V_{idk}$ & Input-hidden parameters\tabularnewline
$\hb$ & A set of hidden variables &  & $w_{k}$ & Hidden bias parameter\tabularnewline
$Z(\cdot)$ & Normalising function &  & $a_{im}$ & Activation indicator\tabularnewline
$\succ$,$\succeq$,$\triangleright$ & Ordinal relations &  & $\mathbb{S}_{i}$ & Set of categories\tabularnewline
$\simeq$ & Indifference &  & $M_{i}$ & The number of categories\tabularnewline
$N$ & Number of visible units &  & $c_{im}$ & Category member of set $\mathbb{S}_{i}$\tabularnewline
$K$ & Number of hidden units &  & $\delta_{m}[v_{i}]$,$\mathbb{I}[\cdot]$ & Indicator functions\tabularnewline
$P(\cdot)$ & Probability distribution &  & $C$ & Index of a subset of variables\tabularnewline
$E(\cdot)$ & Energy function &  & $\mathcal{L}$ & Data log-likelihood\tabularnewline
\hline 
\end{tabular}
\par\end{centering}

\caption{Notations used in this paper. \label{tab:Notations}}
\end{table}

\subsection{Model Definition}

Denote by $\vb=(\v_{1},\v_{2},...,\v_{N})$ the set of \emph{mixed-variate}
visible variables where each $\v_{i}$ can be one of the following
types: \emph{binary}, \emph{categorical}, \emph{multicategorical},
\emph{continuous}, \emph{ordinal} or \emph{category-ranked}. Let $\vb_{disc}$
be the joint set of discrete elements and $\vb_{cont}$ be the continuous
set, and thus $\vb=(\vb_{disc},\vb_{cont})$. Denoting by $\hb=(h_{1},h_{2},...,h_{K})\in\{0,1\}^{K}$
the hidden variables, the model distribution of $\MODEL$ is defined
as
\begin{equation}
P(\vb,\hb)=\frac{1}{Z}\exp\{-E(\vb,\hb)\},\label{eq:model-def}
\end{equation}
where $E(\vb,\hb)$ is the model energy, $Z$ is the normalisation
constant. The model energy is further decomposed into a sum of singleton
and pairwise energies:
\[
E(\vb,\hb)=\sum_{i=1}^{N}E_{i}(\v_{i})+\sum_{k=1}^{K}E_{k}(\h_{k})+\sum_{i=1}^{N}\sum_{k=1}^{K}E_{ik}(\v_{i},\h_{k}),
\]
where $E_{i}(\v_{i})$ depends only on the $i$-th visible unit, $E_{k}(\h_{k})$
on the $k$-th hidden unit, and $E_{ik}(\v_{i},\h_{k})$ on the interaction
between the $i$-th visible and $k$-hidden units. The $\MODEL$ is
thus a $2$-layer mixed-variate Markov random field with pairwise
connectivity across layers.

For the distribution in Eq.~(\ref{eq:model-def}) to be properly
specified, we need to keep the normalisation constant finite. In other
words, the following integration

\[
Z=\int_{\vb_{cont}}\left(\sum_{\vb_{disc}}\sum_{\hb}\exp\{-E(\vb_{disc},\vb_{cont},\hb)\}\right)d(\vb_{cont})
\]
must be bounded from above. One way is to choose appropriate continuous
variable types with bounded moments, e.g., Gaussian. Another way is
to explicitly bound the continuous variables to some finite ball,
i.e., $\left\Vert \vb_{cont}\right\Vert \le R$.

In our $\MODEL$, we further assume that the energies have the following
form:
\begin{equation}
E_{i}(\v_{i})=-G_{i}(\v_{i});\quad\quad\quad\quad E_{k}(h_{k})=-w_{k}h_{k};\quad\quad\quad\quad E_{ik}(\v_{i},\h_{k})=-H_{ik}(v_{i})h_{k},\label{eq:energies}
\end{equation}
where $w_{k}$ is the bias parameter for the $k$-th hidden unit,
and $G_{i}(v_{i})$ and $H_{ik}(v_{i})$ are functions to be specified
for each data type. An important consequence of this energy decomposition
is the factorisation of the \emph{posterior}:

\begin{eqnarray}
P(\hb\mid\vb) & = & \prod_{k}P(h_{k}\mid\vb);\quad\quad\quad\quad P(h_{k}^{1}\mid\vb)=\frac{1}{1+\exp\{-w_{k}-\sum_{i}H_{ik}(\v_{i})\}},\label{eq:posterior}
\end{eqnarray}
where $h_{k}^{1}$ denotes the assignment $h_{k}=1$. This posterior
is efficient to evaluate, and thus the vector $\left(P(h_{k}^{1}\mid\vb),P(h_{k}^{2}\mid\vb),...,P(h_{k}^{K}\mid\vb)\right)$
can be used as extracted features for mixed-variate input $\vb$.

Similarly, the \emph{data model} $P(\vb|\hb)$ has the following factorisation

\begin{equation}
P(\vb\mid\hb)=\prod_{i}P_{i}(\v_{i}\mid\hb);\quad\quad\quad\quad P_{i}(\v_{i}\mid\hb)=\frac{1}{Z(\hb)}\exp\{G_{i}(\v_{i})+\sum_{k}H_{ik}(\v_{i})h_{k}\},\label{eq:data-model}
\end{equation}
where $Z(\hb)=\sum_{\v_{i}}\exp\{G_{i}(\v_{i})+\sum_{k}H_{ik}(\v_{i})h_{k}\}$
if $\v_{i}$ is discrete and $Z(\hb)=\int_{\v_{i}}\exp\{G_{i}(\v_{i})+\sum_{k}H_{ik}(\v_{i})h_{k}\}d(\v_{i})$
if $\v_{i}$ is continuous, assuming that the integration exists.
Note that we deliberately use the subscript index $i$ in $P_{i}(\cdot\mid\hb)$
to emphasize the heterogeneous nature of the input variables.

\subsection{Type-specific Data Models}

We now specify $P_{i}(\v_{i}|\mathbf{h})$ in Eq.~(\ref{eq:data-model}),
or equivalently, the functionals $G_{i}(\v_{i})$ and $H_{ik}(\v_{i})$.
Denote by $\mathbb{S}_{i}=(c_{i1},c_{i2},...,c_{iM_{i}})$ the set
of categories in the case of discrete variables. In this section,
for continuous types, we limit to Gaussian variables as they are the
by far the most common. Interested readers are referred to \cite{le2011learning} for Beta variables
in the context of image modelling. The data model and related functionals for
binary, Gaussian and categorical data types are well-known, and thus we provide
a summary here:

\begin{flushleft}
\begin{tabular}{lccc}
 & $G_{i}(\v_{i})$ & $H_{ik}(\v_{i})$ & $P_{i}(\v_{i}|\hb)$\tabularnewline
\hline 
\noalign{\vskip0.2cm}
--Binary & $U_{i}v_{i}$ & $V_{ik}v_{i}$ & $\frac{\exp\{U_{i}v_{i}+\sum_{k}V_{ik}h_{k}v_{i}\}}{1+\exp\{U_{i}+\sum_{k}V_{ik}h_{k}\}}$\tabularnewline
\noalign{\vskip0.2cm}
--Gaussian & $-\nicefrac{\v_{i}^{2}}{2\sigma_{i}^{2}}+U_{i}v_{i}$ & $V_{ik}v_{i}$ & $\mathcal{N}\left(\sigma_{i}^{2}\left(U_{i}+\sum_{k}V_{ik}h_k\right);\sigma_{i}\right)$\tabularnewline
\noalign{\vskip0.2cm}
--Categorical & $\sum_{m}U_{im}\delta_{m}[v_{i}]$ & $\sum_{m,k}V_{imk}\delta_{m}[v_{i}]$ & $\frac{\exp\{\sum_{m}U_{im}\delta_{m}[v_{i}]+\sum_{m,k}V_{imk}\delta_{m}[v_{i}]h_k\}}{\sum_{l}\exp\{U_{il}+\sum_{k}V_{ilk}h_{k}\}}$\tabularnewline[0.4cm]
\end{tabular}\\
where $m=1,2,...,M_{i}$; $U_{i},V_{ik},U_{im},V_{imk}$ are model
parameters; and $\delta_{m}[v_{i}]=1$ if $v_{i}=c_{im}$ and $0$
otherwise.
\par\end{flushleft}

The cases of multicategorical, ordinal and category-ranking variables
are, however, much more involved, and thus some further simplification
may be necessary. In what follows, we describe the specification details
for these three cases.

\subsubsection{Multicategorical Variables \label{sub:Multi-categorical-Variables}}

An assignment to a multicategorical variable has the form of a subset
from a set of categories. For example, a person may be interested
in \textsf{games} and \textsf{music} from a set of offers: \textsf{games},
\textsf{sports}, \textsf{music}, and \textsf{photography}. More formally,
let $\mathbb{S}_{i}$ be the set of categories for the $i$-th variable,
and $\mathcal{P}_{i}=2^{\mathbb{S}_{i}}$ be the power set of $\mathbb{S}_{i}$
(the set of all possible subsets of $\mathbb{S}_{i}$). Each variable
assignment consists of a non-empty element of $\mathcal{P}_{i}$,
i.e. $\v_{i}\in\left\{ \mathcal{P}_{i}\backslash\emptyset\right\} $.
Since there are $2^{M_{i}}-1$ possible ways to select a non-empty
subset, directly enumerating $P_{i}(\v_{i}|\hb)$ proves to be highly
difficult even for moderate sets. To handle this state explosion,
we first assign each category $c_{im}$ with a binary indicator $a_{im}\in\{0,1\}$
to indicate whether the $m$-th category is active, that is $\v_{i}=\left(a_{i1},a_{i2},...,a_{iM_{i}}\right)$.
We then assume the following factorisation:
\begin{equation}
P_{i}(\v_{i}|\hb)=\prod_{m=1}^{M_{i}}P_{i}(a_{im}|\hb).\label{eq:multilabel-data-model}
\end{equation}
Note that this does not says that binary indicators are independent
in their own right but given the knowledge of the hidden variables
$\hb$. Since they hidden variables are never observed, binary indicators
are therefore interdependent. Now, the probability for activating
a binary indicator is defined as

\begin{equation}
P_{i}(a_{im}=1|\hb)=\frac{1}{1+\exp(-U_{im}-\sum_{k}V_{imk}\h_{k})}.\label{eq:multicat-model}
\end{equation}

Note that this specification is equivalent to the following decomposition
of the functionals $G_{i}(v_{i})$ and $H_{ik}(\v_{i})$ in Eq.~(\ref{eq:energies}):
\begin{align*}
G_{i}(v_{i}) & =\sum_{m=1}^{M_{i}}U_{im}a_{im};\quad\quad\quad\quad H_{ik}(\v_{i})=\sum_{m=1}^{M_{i}}V_{imk}a_{im}.
\end{align*}

\subsubsection{Ordinal Variables}

An ordinal variable receives individual values from an ordinal set
$S_{i}=\{c_{i1}\prec c_{i2}\prec...,\prec c_{iM_{i}}\}$ where $\prec$
denotes the order in some sense. For example, $c_{im}$ can be a numerical
rating from a review, or it can be sentimental expression such as
\textsf{love}, \textsf{neutral} and \textsf{hate}. There are two straightforward
ways to treat an ordinal variable: (i) one is simply ignoring the
order, and considering it as a multinomial variable, and (ii) another
way is to convert the ordinal expression into some numerical scale,
for example, $\{-1,0,+1\}$ for the triple \{\textsf{love}\textsf{\emph{,}}\textsf{neutral}\textsf{\emph{,}}\textsf{hate}\}
and then proceed as if it is a continuous variable. However, in the
first treatment, substantial ordinal information is lost, and in the
second treatment, there is no satisfying interpretation using numbers.

In this paper, we adapt the Stereotype Ordered Regression Model (SORM)
by \cite{anderson1984rao}. More specifically, the SORM defines the
conditional distribution as follows
\[
P(v_{i}=m\mid\hb)=\frac{\exp\{U_{im}+\sum_{d=1}^{D}\sum_{k=1}^{K}V_{idk}\phi_{id}(m)h_{k}\}}{\sum_{l}\exp\{U_{il}+\sum_{d=1}^{D}\sum_{k=1}^{K}V_{idk}\phi_{id}(l)h_{k}\}}
\]
where $U_{im},V_{idk}$ are free parameters, $D\le M_{i}$ is the
dimensionality of the ordinal variable%
\footnote{This should not be confused with the dimensionality of the whole data
$\vb$.%
} $\v_{i}$, and $\phi_{id}(m)$ is the monotonically increasing function
of $m$: 
\[
\phi_{id}(1)<\phi_{id}(2)<...<\phi_{id}(M_{i})
\]

A shortcoming of this setting is that when $\hb=\boldsymbol{0}$,
the model reduces to the standard multiclass logistic, effectively
removing the ordinal property. To deal with this, we propose to make
the input bias parameters order dependent:
\begin{equation}
P(v_{i}=m\mid\hb)\propto\exp\left\{ \sum_{d=1}^{D}\phi_{id}(m)\left(U_{id}+\sum_{k=1}^{K}V_{idk}h_{k}\right)\right\} \label{eq:ordinal-model}
\end{equation}
where $U_{id}$ is the newly introduced parameter. Here we choose
$D=M_{i}$, and $\phi_{id}(m)=\nicefrac{\left(m-d\right)}{\left(M_{i}-1\right)}$.

\subsubsection{Category-ranking Variables \label{sub:Category-ranking-Variables}}

In category ranking, a variable assignment has the form of a ranked
list of a set of categories. For example, from a set of offers namely
\textsf{games}, \textsf{sports}, \textsf{music}, and \textsf{photography},
a person may express their preferences in a particular decreasing
order: \textsf{sports} $\succ$ \textsf{music} $\succ$ \textsf{games} $\succ$ \textsf{photography}.
Sometimes, they may like sports and music equally, creating a situation
known as \emph{ties} in ranking, or \emph{indifference} in preference.
When there are no ties, we can say that the rank is \emph{complete}.

More formally, from a set of categories $\mathbb{S}_{i}=\{c_{i1},c_{i2},...,c_{iM_{i}}\}$,
a variable assignment without ties is then a permutation of elements
of $\mathbb{S}_{i}$. Thus, there are $M_{i}!$ possible complete
rank assignments. When we allow ties to occur, however, the number
of possible assignments is extremely large. To see how, let us group
categories of the same rank into a partition. Orders within a partition
are not important, but orders between partitions are. Thus, the problem
of rank assignment turns out to be choosing from a set of all possible
schemes for partitioning and ordering a set. The number of such schemes
is known in combinatorics as the \emph{Fubini's number} \cite[pp. 396--397]{muresan2008concrete},
which is extremely large even for small sets. For example, $\mbox{Fubini}\left(1\right)=1$,
$\mbox{Fubini}\left(3\right)=13$, $\mbox{Fubini}\left(5\right)=541$
and $\mbox{Fubini}\left(10\right)=102,247,563$. Directly modelling
ranking with ties proves to be intractable. 

We thus resort to approximate methods. One way is to model just pairwise
comparisons: we treat each pair of categories separately \emph{when
conditioned on the hidden layer}. More formally, denote by $c_{il}\succ c_{im}$
the preference of category $c_{il}$ over $c_{im}$, and by $c_{il}\simeq c_{im}$
the indifference. We replace the data model $P_{i}(\v_{i}|\hb)$ with
a product of pairwise comparisons $\prod_{l}\prod_{m>l}P_{i}(c_{il}\triangleright c_{im}|\hb)$,
where $\triangleright$ denotes preference relations (i.e., $\succ$,
$\prec$ or $\simeq$). This effectively translates the original problem
with Fubini's number complexity to $M_{i}(M_{i}-1)/2$ pairwise sub-problems,
each of which has only three preference choices. The drawback is that
this relaxation loses the guarantee of \emph{transitivity }(i.e.,
$c_{il}\succeq c_{im}$ and $c_{im}\succeq c_{in}$ would entail $c_{il}\succeq c_{in}$,
where $\succeq$ means \emph{better} or \emph{equal-to}). The hope
is that the hidden layer is rich enough to absorb this property, that
is, the probability of preserving the transitivity is sufficiently
high.

Now it remains to specify $P_{i}(c_{il}\triangleright c_{im}|\hb)$
in details. In particular, we adapt the Davidson's model \cite{davidson1970extending}
of pairwise comparison:
\begin{eqnarray}
P_{i}(c_{il}\succ c_{im}|\hb) & = & \frac{1}{Z_{iml}(\hb)}\varphi(c_{il},\hb)\nonumber \\
P_{i}(c_{il}\simeq c_{im}|\hb) & = & \frac{1}{Z_{iml}(\hb)}\gamma\sqrt{\varphi(c_{il},\hb)\varphi(c_{im},\hb)}\label{eq:label-rank-model}\\
P_{i}(c_{il}\prec c_{im}|\hb) & = & \frac{1}{Z_{iml}(\hb)}\varphi(c_{im},\hb)\nonumber 
\end{eqnarray}
where $Z_{ilm}(\hb)=\varphi(c_{il},\hb)+\varphi(c_{im},\hb)+\gamma\sqrt{\varphi(c_{il},\hb)\varphi(c_{im},\hb)}$,
$\gamma>0$ is the tie parameter, and 

\[
\varphi(c_{im},\hb)=\exp\{\frac{1}{M_{i}}(U_{im}+\sum_{k}V_{imk}h_{k})\}.
\]
The term $\nicefrac{1}{M_{i}}$ normalises the occurrence frequency
of a category in the model energy, leading to better numerical stability.

%% file: learning.tex
In this paper, we consider two applications of the $\MODEL$: \emph{estimating
data distribution} and \emph{learning predictive model}s. Estimating
data distribution is to learn a generative model that generates
the visible data. This can be useful in many other applications including
dimensionality reduction, feature extraction, and data completion.
On the other hand, a predictive model is a classification (or regression)
tool that predicts an output given the input co-variates.

\subsection{Parameter Learning \label{sub:Learning}}

We now present parameter estimation for $\{w_{k},U_{i},U_{im},V_{ik},V_{imk}\}$,
which clearly depend on the specific applications.

\subsubsection{Estimating Data Distribution \label{sub:Estimating-Data-Distribution}}

The problem of estimating a distribution from data is typically performed
by maximising the data likelihood $\mathcal{L}_{1}=\sum_{\vb}\tilde{P}(\vb)\log P(\vb)$,
where $\tilde{P}(\vb)$ denotes the empirical distribution of the
visible variables, and $P(\vb)=\sum_{\hb}P(\vb,\hb)$ is the model
distribution. Since the $\MODEL$ belongs to the exponential family,
the gradient of $\mathcal{L}_{1}$ with respect to parameters takes
the form of difference of expectations. For example, in the case of
binary variables, the gradient reads
\[
\frac{\partial\mathcal{L}_{1}}{\partial V_{ik}}=\left\langle v_{i}h_{k}\right\rangle _{\tilde{P}(v_{i},h_{k})}-\left\langle v_{i}h_{k}\right\rangle _{P(v_{i},h_{k})}
\]
where $\tilde{P}(h_{k},v_{i})=P(h_{k}|\vb)\tilde{P}(v_{i})$ is the
empirical distribution, and $P(h_{k},v_{i})=P(h_{k}|\vb)P(v_{i})$
the model distribution. Due to space constraint, we omit the derivation
details here. 

The empirical expectation $\left\langle v_{i}h_{k}\right\rangle _{\tilde{P}(v_{i},h_{k})}$
is easy to estimate due to the factorisation in Eq.~(\ref{eq:posterior}).
However, the model expectation $\left\langle v_{i}h_{k}\right\rangle _{P(v_{i},h_{k})}$
is intractable to evaluate exactly, and thus we must resort to approximate
methods. Due to the factorisations in Eqs.~(\ref{eq:posterior},\ref{eq:data-model}),
Markov Chain Monte Carlo samplers are efficient to run. More specifically,
the sampler is alternating between $\left\{ \widehat{h}_{k}\sim P(h_{k}|\vb)\right\} _{k=1}^{K}$
and $\left\{ \widehat{\v}_{i}\sim P(\v_{i}|\hb)\right\} _{i=1}^{N}$.
Note that in the case of multicategorical variables, make use of the
factorisation in Eq.~(\ref{eq:multilabel-data-model}) and sample
$\{a_{im}\}_{m=1}^{M_{i}}$ simultaneously. On the other hand, in
the case of category-ranked variables, we do not sample directly from
$P(\v_{i}|\hb)$ but from its relaxation $\left\{ P_{i}(c_{il}\triangleright c_{im}|\hb)\right\} _{l,m>l}$
- which have the form of multinomial distributions. To speed up, we
follow the method of Contrastive Divergence (CD) \cite{Hinton02},
in which the MCMC is restarted from the observed data $\vb$ and stopped
after just a few steps for every parameter update. This has been known
to introduce bias to the model estimate, but it is often fast and
effective for many applications.

For the data completion application, in the data we observed only
some variables and others are missing. There are two ways to handle
a missing variable during training time: one is to treat it as hidden,
and the other is to ignore it. In this paper, we follows the latter
for simplicity and efficiency, especially when the data is highly
sparse\footnote{Ignoring missing data may be inadequate if 
the missing patterns are not at random. However, treating missing data as 
zero observations (e.g., in the case of binary variables) may not be accurate 
either since it may introduce bias to the data marginals.}.

\subsubsection{Learning Predictive Models \label{sub:Learning-Predictive-Models}}

In our $\MODEL$, a predictive task can be represented by an output
variable conditioned on input variables. Denote by $v_{i}$ the $i$-th
output variable, and $\vb_{\neg i}$ the set of input variables, that
is, $\vb=(\v_{i},\vb_{\neg i})$. The learning problem is translated
into estimating the conditional distribution $P(\v_{i}\mid\vb_{\neg i})$. 

There are three general ways to learn a predictive model. The \emph{generative}
method first learns the joint distribution $P(\v_{i},\vb_{\neg i})$
as in the problem of estimating data distribution. The \emph{discriminative}
method, on the other hand, effectively ignores $P(\vb_{\neg i})$
and concentrates only on $P(\v_{i}\mid\vb_{\neg i})$. In the latter,
we typically maximise the conditional likelihood $\mathcal{L}_{2}=\sum_{\v_{i}}\sum_{\vb_{\neg i}}\tilde{P}(\v_{i},\vb_{\neg i})\log P(\vb_{i}\mid\vb_{\neg i})$.
This problem is inherently easier than the former because we do not
have to make inference about $\vb_{\neg i}$. The learning strategy
is almost identical to that of the generative counterpart, except
that we \emph{clamp} the input variables $\vb_{\neg i}$ to their
observed values. For tasks whose size of the output space is small
(e.g., standard binary, ordinal, categorical variables) we can perform
exact evaluations and use any non-linear optimisation methods for
parameter estimation. The conditional distribution $P(\v_{i}\mid\vb_{\neg i})$
can be computed as in Eq.~(\ref{eq:predictive-distribution}). We
omit the likelihood gradient here for space limitation.

It is often argued that the discriminative method is more preferable
since there is no waste of effort in learning $P(\vb_{\neg i})$,
which we do not need at test time. In our setting, however, learning
$P(\vb_{\neg i})$ may yield a more faithful representation%
\footnote{As we do not need labels to learn $P(\vb_{\neg i})$, this is actually
a form of \emph{semi-supervised learning}.%
} of the data through the posterior $P(\hb\mid\vb_{\neg i})$. This
suggests a third, \emph{hybrid} method: combining the generative and
discriminative objectives. One way is to optimise a hybrid likelihood:

\[
\mathcal{L}_{3}=\lambda\sum_{\vb_{\neg i}}\tilde{P}(\vb_{\neg i})\log P(\vb_{\neg i})+(1-\lambda)\sum_{\vb_{i}}\sum_{\vb_{\neg i}}\tilde{P}(\vb_{i},\vb_{\neg i})\log P(\vb_{i}\mid\vb_{\neg i}),
\]
where $\lambda\in(0,1)$ is the hyper-parameter controlling the relative
contribution of generative and discriminative components. Another
way is to use a \textbf{$2$}-stage procedure: first we \emph{pre-train}
the model $P(\vb_{\neg i})$ in an unsupervised manner, and then \emph{fine-tune}
the predictive model%
\footnote{We can also avoid tuning parameters associated with $\vb_{\neg i}$
by using the posteriors as features and learn $P\left(\v_{i}\mid\hat{\hb}\right)$,
where $\hat{\h}_{k}=P\left(h_{k}^{1}\mid\vb_{\neg i}\right).$%
} $P(\vb_{i}\mid\vb_{\neg i})$.

\subsection{Prediction \label{sub:Prediction}}

Once the model has been learnt, we are ready to perform prediction.
We study two predictive applications: completing missing data, and
output labels in predictive modelling. The former leads to the inference
of $P(\vb_{C}\mid\vb_{\neg C})$, where $\vb_{\neg C}$ is the set
of observed variables, and $\vb_{C}$ is the set of unseen variables
to be predicted. Ideally, we should predict all unseen variables simultaneously
but the inference is likely to be difficult. Thus, we resort to estimating
$P(v_{i}|\vb_{\neg C})$, for $i\in C$. The prediction application
requires the estimation of $P(v_{i}|\vb_{\neg i})$, which is clearly
a special case of $P(v_{i}|\vb_{\neg C})$, i.e., when $C=\{i\}$.
The output is predicted as follows
\begin{align}
\hat{\v}_{i} & =\arg\max_{\v_{i}}P(v_{i}|\vb_{\neg C})=\arg\max_{\v_{i}}\sum_{\hb}P(v_{i},\hb|\vb_{\neg C});\quad\mbox{where}\label{eq:prediction}\\
P(v_{i},\hb|\vb_{\neg C}) & =\frac{1}{Z(\vb_{\neg C},\hb)}\exp\left\{ G_{i}(v_{i})+\sum_{k}w_{k}h_{k}+\sum_{j\in\{\neg C,i\}}\sum_{k}H_{jk}(v_{j})h_{k}\right\} ,\label{eq:output-hidden-prob}
\end{align}
where $Z(\vb_{\neg C})$ is the normalising constant. Noting that
$\h_{k}\in\{0,1\}$, the computation of $P(v_{i}|\vb_{\neg C})$ can
be simplified as
\begin{align}
P(v_{i}|\vb_{\neg C}) & =\frac{1}{Z(\vb_{\neg C})}\exp\{G_{i}(v_{i})\}\prod_{k}\left[1+\frac{\exp\{H_{ik}(v_{i})\}}{1/P(h_{k}^{1}|\vb_{\neg C})-1}\right]\label{eq:predictive-distribution}
\end{align}
where $P(h_{k}^{1}|\vb_{\neg C})$ is computed using Eq.~(\ref{eq:posterior})
as
\[
P(h_{k}^{1}|\vb_{\neg C})=\frac{1}{1+\exp\{-w_{k}-\sum_{j\in\neg C}H_{jk}(v_{j})\}}.
\]

For the cases of binary, categorical and ordinal outputs, the estimation
in Eq.~(\ref{eq:prediction}) is straightforward using Eq.~(\ref{eq:predictive-distribution}).
However, for other output types, suitable simplification must be made:
\begin{itemize}
\item For multicategorical and category-ranking variables, we do not enumerate
over all possible assignments of $v_{i}$, but rather in an indirect
manner:

\begin{itemize}
\item For multiple categories (Section~\ref{sub:Multi-categorical-Variables}),
we first estimate $\left\{ P_{i}(a_{im}=1|\vb_{\neg i})\right\} _{m=1}^{M_{i}}$
and then output $a_{im}=1$ if $P_{i}(a_{im}=1|\vb_{\neg i})\ge\nu$
for some threshold%
\footnote{Raising the threshold typically leads to better precision at the expense
of recall. Typically we choose $\nu=0.5$ when there is no preference
over recall nor precision.%
} $\nu\in(0,1)$. 
\item For category-ranking (Section~\ref{sub:Category-ranking-Variables}),
we first estimate $\left\{ P_{i}(c_{il}\succ c_{im}|\vb_{\neg i})\right\} _{l,m>l}$.
The complete ranking over the set $\{c_{i1},c_{i2},...,c_{iM_{i}}\}$
can be obtained by aggregating over probability pairwise relations.
For example, the score for $c_{im}$ can be estimated as $s(c_{im})=\sum_{l\ne m}P_{i}(c_{im}\succ c_{il}|\vb_{\neg i})$,
which can be used for sorting categories%
\footnote{Note that we do not estimate the event of ties during prediction.%
}. 
\end{itemize}
\item For continuous variables, the problem leads to a non-trivial nonlinear
optimisation: even for the case of Gaussian variables, $P(v_{i}|\vb_{\neg C})$
in Eq.~(\ref{eq:predictive-distribution}) is no longer Gaussian.
For efficiency and simplicity, we can take a \emph{mean-field} approximation
by substituting $\hat{\h}_{k}=P(h_{k}^{1}|\vb_{\neg C})$ for $\h_{k}$.
For example, in the case of Gaussian outputs, we then obtain a simplified
expression for $P(v_{i}|\vb_{\neg C})$: 
\[
P(v_{i}|\vb_{\neg C})\propto\exp\left\{ -\frac{\v_{i}^{2}}{2\sigma_{i}^{2}}+U_{i}\v_{i}+\sum_{k}V_{ik}v_{i}\hat{\h}_{k}\right\} ,
\]
which is also a Gaussian. Thus the optimal value is the mean itself:
$\hat{\v}_{i}=\sigma_{i}^{2}\left(U_{i}+\sum_{k}V_{ik}\hat{\h}_{k}\right).$
Details of the mean-field approximation is presented in Appendix~\ref{sub:Mean-field-Approximation}.\end{itemize}

%% file: experiments.tex
\subsection{Setting}

In this experiment, we run the $\MODEL$ on a large-scale survey of
the general world opinion, which was published by the Pew Global Attitudes
Project%
\footnote{http://pewglobal.org/datasets/%
} in the summer of $2002$. The survey was based on interviewing with
people in $44$ countries in the period of $2001$--$2002$. Some
sample questions are listed in Appendix~\ref{sec:Sample-Questions}.
After some pre-processing, we obtain a dataset of $38,263$ people,
each of whom provides answers to a subset of $189$ questions over
multiple topics ranging from globalization, democracy to terrorism.
Many answers are deliberately left empty because it may be inappropriate
to ask certain type of questions in a certain area or ethnic group.
Of all answers, $43$ are binary, $12$ are categorical, $3$ are
multicategorical, $125$ are ordinal, $2$ are category-ranking, and
$3$ are continuous. To suppress the scale difference in continuous
responses, we normalise them to zeros means and 
unit variances~\footnote{It may be desirable to learn the variance structure, but we keep it simple 
by fixing to unit variance. For more sophisticated variance learning, we refer to a recent paper \cite{le2011learning}
for more details.}.

We evaluate each data type separately. In particular, let $u$ be
the user index, $\hat{\v}{}_{i}$ be the predicted value of the $i$-th
variable, and $N_{t}$ is the number of variables of type $t$ in
the test data, we compute the prediction errors as follows:

\begin{tabular}{ll}
\noalign{\vskip0.2cm}
--Binary & :~~$\frac{1}{N_{bin}}\sum_{u}\sum_{i}\mathbb{I}\left[\v_{i}^{(u)}\ne\hat{\v}_{i}^{(u)}\right]$,\tabularnewline[0.2cm]
\noalign{\vskip0.2cm}
--Categorical & :~~$\frac{1}{N_{cat}}\sum_{u}\sum_{i}\mathbb{I}\left[\v_{i}^{(u)}\ne\hat{\v}_{i}^{(u)}\right]$,\tabularnewline[0.2cm]
\noalign{\vskip0.2cm}
--Multicategorical & :~~$1-\nicefrac{2\mathrm{R}\mathrm{P}}{\left(\mathrm{R}+\mathrm{P}\right)}$,\tabularnewline[0.2cm]
\noalign{\vskip0.2cm}
--Continuous & :~~$\sqrt{\frac{1}{D_{cont}}\sum_{u}\sum_{i}\left(\v_{i}^{(u)}-\hat{\v}_{i}^{(u)}\right)^{2}}$,\tabularnewline[0.2cm]
\noalign{\vskip0.2cm}
--Ordinal & :~~$\frac{1}{N_{ord}}\sum_{u}\sum_{i}\frac{1}{M_{i}-1}\left|\v_{i}^{(u)}-\hat{\v}_{i}^{(u)}\right|$,\tabularnewline[0.2cm]
\noalign{\vskip0.2cm}
--Category-ranking & :~~$\frac{1}{D_{rank}}\sum_{u}\sum_{i}\frac{2}{M_{i}(M_{i}-1)}\sum_{l,m>l}\mathbb{I}\left[(\pi_{il}^{(u)}-\pi_{im}^{(u)})(\hat{\pi}_{il}^{(u)}-\hat{\pi}_{im}^{(u)})<0\right]$,\tabularnewline[0.2cm]
\end{tabular}\\
where $\mathbb{I}\left[\cdot\right]$ is the identity function, $\pi_{im}\in\{1,2,...,M_{i}\}$
is the rank of the $m$-th category of the $i$-th variable, $\mathrm{R}$
is the recall rate and $\mathrm{P}$ is the precision. The recall
and precision are defined as:
\[
\mathrm{R}=\frac{\sum_{u}\sum_{i}\frac{1}{M_{i}}\sum_{m=1}^{M_{i}}\mathbb{I}\left[a_{im}^{(u)}=\hat{a}_{im}^{(u)}\right]}{\sum_{u}\sum_{i}\frac{1}{M_{i}}\sum_{m=1}^{M_{i}}a_{im}^{(u)}};\quad\quad\quad\mathrm{P}=\frac{\sum_{u}\sum_{i}\frac{1}{M_{i}}\sum_{m=1}^{M_{i}}\mathbb{I}\left[a_{im}^{(u)}=\hat{a}_{im}^{(u)}\right]}{\sum_{u}\sum_{i}\frac{1}{M_{i}}\sum_{m=1}^{M_{i}}\hat{a}_{im}^{(u)}},
\]
where $a_{im}\in\{0,1\}$ is the $m$-th component of the $i$-th
multicategorical variable. Note that the summation over $i$ for each
type only consists of relevant variables.

To create baselines, we use the $\MODEL$ without the hidden layer,
i.e., by assuming that variables are independent\footnote{To the best of our 
knowledge, there has been no totally comparable work addressing the issues we study
in this paper. Existing survey analysis methods are suitable
for individual tasks such as measuring
pairwise correlation among variables, or building individual 
regression models where complex co-variates are coded into binary variables.}.

\subsection{Results}

\subsubsection{Feature Extraction and Visualisation}

\begin{table*}
\begin{centering}
\begin{tabular}{|l|c|ccccc|}
\hline 
 & Baseline  & $K=20$ & $K=50$ & $K=100$ & $K=200$ & $K=500$\tabularnewline
\hline 
Binary  & 32.9 & 23.6 & 20.1 & 16.3 & 13.2 & 9.8\tabularnewline
\hline 
Categorical  & 52.3 & 29.8 & 22.0 & 17.0 & 13.2 & 7.1\tabularnewline
\hline 
Multicategorical  & 49.6 & 46.6 & 42.2 & 36.9 & 29.2 & 23.8\tabularnewline
\hline 
Continuous({*})  & 100.0 & 89.3 & 84.1 & 78.4 & 69.5 & 65.5\tabularnewline
\hline 
Ordinal  & 25.2 & 19.5 & 16.2 & 13.5 & 10.9 & 7.7\tabularnewline
\hline 
Category ranking  & 19.3 & 11.7 & 6.0 & 5.0 & 3.2 & 2.3\tabularnewline
\hline 
\end{tabular}
\par\end{centering}

\caption{Error rates (\%) when reconstructing data from posteriors. The baseline
is essentially the $\MODEL$ without hidden layer (i.e., assuming
variables are independent). ({*}) The continuous variables have been
normalised to account for different scales between items, thus the
baseline error will be $1$ (i.e., the unit variance).\label{tab:Reconstruction-error-rates}}
\end{table*}

Recall that our $\MODEL$ can be used as a feature extraction tool
through the posterior projection. The projection converts a multimodal
input into a real-valued vector of the form $\hat{\hb}=\left(\hat{\h}_{1},\hat{\h}_{2},...,\hat{\h}_{K}\right)$,
where $\hat{\h}_{k}=P\left(\h_{k}=1\mid\vb\right)$. Clearly, numerical
vectors are much easier to process further than the original data,
and in fact the vectorial form is required for the majority of modern data handling
tools (e.g., for transformation, clustering, comparison and visualisation).
To evaluate the faithfulness of the new representation, we reconstruct
the original data using $\hat{\v}_{i}=\arg\max_{\v_{i}}P\left(\v_{i}\mid\hat{\hb}\right)$,
that is, in Eq.~(\ref{eq:data-model}), the binary vector $\hb$
is replaced by $\hat{\hb}$. The use of $P\left(\v_{i}\mid\hat{\hb}\right)$
can be reasoned through the mean-field approximation framework presented
in Appendix~\ref{sub:Mean-field-Approximation}. Table~\ref{tab:Reconstruction-error-rates}
presents the reconstruction results. The trends are not surprising:
with more hidden units, the model becomes more flexible and accurate
in capturing the data content.

\begin{figure}
\begin{centering}
\includegraphics[height=0.5\linewidth]{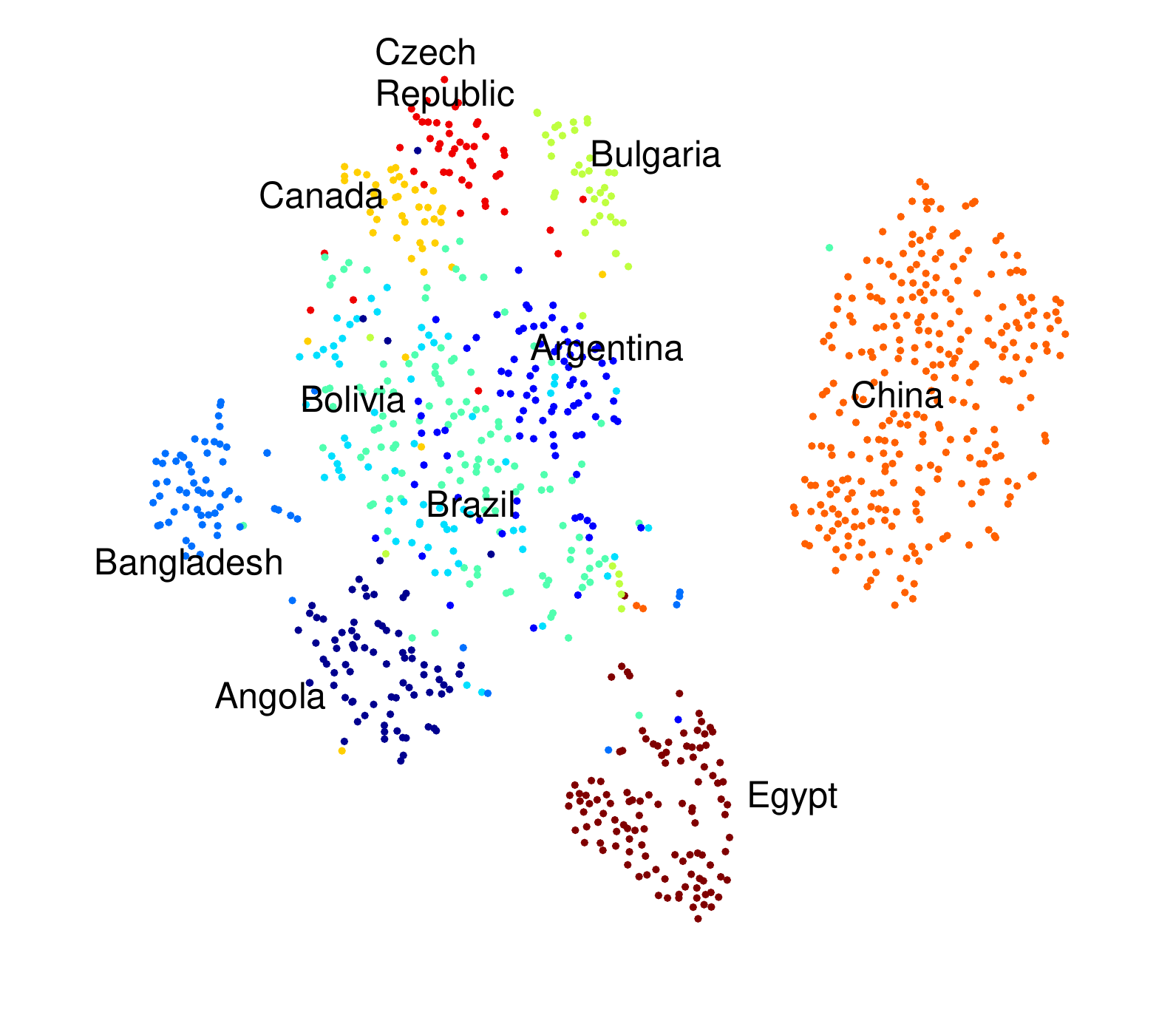} 
\par\end{centering}

\caption{t-SNE projection of posteriors ($K=50$) with country information
removed. Each point is a person from one of the $10$ countries: Angola,
Argentina, Bangladesh, Bolivia, Brazil, Bulgaria, Canada, China, Czech
Republic, and Egypt. Each colour represents a country. Best viewed
in colour. \label{fig:t-SNE-projection}}
\end{figure}

For visualisation, we first learn our $\MODEL$ (with $K=50$ hidden
units) using randomly chosen $3,830$ users, with the country information
removed. Then we use the t-SNE \cite{van2008visualizing} to project
the posteriors further into 2D. Figure~\ref{fig:t-SNE-projection}
shows the distribution of people's opinions in $10$ countries (Angola,
Argentina, Bangladesh, Bolivia, Brazil, Bulgaria, Canada, China, Czech
Republic, and Egypt). It is interesting to see how opinions cluster
geographically and culturally: Europe \& North America (Bulgaria,
Canada \& Czech Republic), South America (Argentina, Bolivia, Brazil),
East Asia (China), South Asia (Bangladesh), North Africa (Egypt) and
South Africa (Angola).

\subsubsection{Data Completion}

In this task, we need to fill missing answers for each survey response.
Missing answers are common in real survey data because the respondents
may forget to answer or simply ignore the questions. We create an
evaluation test by randomly removing a portion $\rho\in(0,1)$ of
answers for each person. The $\MODEL$ is then trained on the remaining
answers in a generative fashion (Section~\ref{sub:Estimating-Data-Distribution}).
Missing answers are then predicted as in Section~\ref{sub:Prediction}.
The idea here is that missing answers of a person can be interpolated
from available answers by other persons. This is essentially a multimodal
generalisation of the so-called collaborative filtering problem. Table~\ref{tab:Completion-errors-full}
reports the completion results for a subset of the data.

\begin{table*}
\begin{centering}
\begin{tabular}{|l|c|ccccc|}
\hline 
 & Baseline  & $K=20$ & $K=50$ & $K=100$ & $K=200$ & $K=500$\tabularnewline
\hline 
Binary  & 32.7 & 26.0 & 24.2 & 23.3 & 22.7 & 22.3\tabularnewline
\hline 
Categorical  & 52.1 & 34.3 & 30.0 & 28.2 & 27.5 & 27.1\tabularnewline
\hline 
Multicategorical  & 49.5 & 48.3 & 45.7 & 43.6 & 42.4 & 42.0\tabularnewline
\hline 
Continuous({*})  & 101.6 & 93.5 & 89.9 & 87.9 & 87.3 & 87.9\tabularnewline
\hline 
Ordinal  & 25.1 & 20.7 & 19.3 & 18.6 & 18.2 & 17.9\tabularnewline
\hline 
Category ranking  & 19.3 & 15.4 & 14.7 & 14.2 & 14.1 & 13.9\tabularnewline
\hline 
\end{tabular}
\par\end{centering}

\caption{Completion error rates (\%) $\rho=0.2$ answers missing at random.
({*}) See Table~\ref{tab:Reconstruction-error-rates}.\label{tab:Completion-errors-full} }
\end{table*}

\subsubsection{Learning Predictive Models}

We study six predictive problems, each of which is representative
for a data type. This means six corresponding variables are reserved
as outputs and the rest as input co-variates. The predictive problems
are: (i) satisfaction with the country (\emph{binary}), (ii) country
of origin (\emph{categorical}, of size $44$), (iii) problems facing
the country (\emph{multicategorical}, of size $11$), (iv) age of
the person (\emph{continuou}s), (v) ladder of life (\emph{ordinal},
of size $11$), and (vi) rank of dangers of the world (\emph{category-ranking},
of size $5$). All models are trained discriminatively (see Section~\ref{sub:Learning-Predictive-Models}).
We randomly split the users into a training subset and a testing subset.
The predictive results are presented in Table~\ref{tab:Predictive-errors-full}.
It can be seen that learning predictive models requires far less number
of hidden units than the tasks of reconstruction and completion. This
is because in discriminative training, the hidden layer acts as an
information filter that allows relevant amount of bits passing from
the input to the output. Since there is only one output per prediction
task, the number of required bits, therefore number of hidden units,
is relatively small. In reconstruction and completion, on the other
hand, we need many bits to represent all the available information.

\begin{table*}
\begin{centering}
\begin{tabular}{|l|c|cccccc|}
\hline 
 & Baseline & $K=3$ & $K=5$ & $K=10$ & $K=15$ & $K=20$ & $K=50$\tabularnewline
\hline 
Satisfaction (\emph{bin}.)  & 26.3 & 18.0 & 17.7 & 17.7 & 17.8 & 18.0 & 18.0\tabularnewline
\hline 
Country (\emph{cat}.) & 92.0 & 70.2 & 61.0 & 21.6 & 11.0 & 9.9 & 5.9\tabularnewline
\hline 
Probs. (\emph{multicat}.)  & 49.6 & 47.6 & 41.9 & 39.2 & 38.8 & 39.1 & 39.2\tabularnewline
\hline 
Age (\emph{cont}.{*})  & 99.8 & 67.3 & 67.6 & 66.3 & 66.4 & 65.8 & 66.3\tabularnewline
\hline 
Life ladder (\emph{ord}.) & 16.9 & 12.2 & 12.2 & 11.9 & 11.9 & 12.2 & 11.8\tabularnewline
\hline 
Dangers (\emph{cat.-rank})  & 31.2 & 27.1 & 24.6 & 24.0 & 23.2 & 23.0 & 22.5\tabularnewline
\hline 
\end{tabular}
\par\end{centering}

\caption{Predictive error rates (\%) with $80/20$ train/test split. ({*})
See Table~\ref{tab:Reconstruction-error-rates}. \label{tab:Predictive-errors-full}}
\end{table*}

%% file: appendix.tex
\subsection{Sample Questions \label{sec:Sample-Questions}}
\begin{itemize}
\item \textbf{Q1} (\emph{Ordinal}): How would you describe your day today\textemdash{}has
it been a typical day, a particularly good day, or a particularly
bad day? 
\item \textbf{Q7} (\emph{Binary}): Now thinking about our country, overall,
are you satisfied or dissatisfied with the way things are going in
our country today? 
\item \textbf{Q5} (\emph{Multicategorical}): What do you think is the most
important problem facing you and your family today? \{Economic problems
/ Housing / Health / Children and education/Work/Social relations
/ Transportation / Problems with government / Crime / Terrorism and
war / No problems / Other / Don't know / Refused\} 
\item \textbf{Q10,11} (\emph{Category-ranking}): In your opinion, which
one of these poses the greatest/second greatest threat to the world:
\{the spread of nuclear weapons / religious and ethnic hatred/AIDS
and other infectious diseases / pollution and other environmental
problems / or the growing gap between the rich and poor\}?
\item \textbf{Q74} (\emph{Continuous}): How old were you at your last birthday? 
\item \textbf{Q91} (\emph{Categorical}): Are you currently married or living
with a partner, widowed, divorced, separated, or have you never been
married? 
\end{itemize}

\subsection{Mean-field Approximation \label{sub:Mean-field-Approximation}}

We present here a simplification of $P\left(v_{i}\mid\vb_{\neg C}\right)$
in Eq.~(\ref{eq:predictive-distribution}) using the mean-field approximation.
Recall that $P\left(v_{i}\mid\vb_{\neg C}\right)=\sum_{\hb}P\left(v_{i},\hb\mid\vb_{\neg C}\right)$,
where $P\left(v_{i},\hb\mid\vb_{\neg C}\right)$ is defined in Eq.~(\ref{eq:output-hidden-prob}).
We approximate $P\left(v_{i},\hb\mid\vb_{\neg C}\right)$ by a fully
factorised distribution 
\[
Q\left(v_{i},\hb\mid\vb_{\neg C}\right)=Q\left(\v_{i}\mid\vb_{\neg C}\right)\prod_{k}Q\left(\h_{k}\mid\vb_{\neg C}\right).
\]
The approximate distribution $Q\left(v_{i},\hb\mid\vb_{\neg C}\right)$
is obtained by minimising the Kullback-Leibler divergence 
\[
\mathcal{D}_{KL}\left(Q\left(v_{i},\hb\mid\vb_{\neg C}\right)\parallel P\left(v_{i},\hb\mid\vb_{\neg C}\right)\right)=\sum_{\v_{i}}\sum_{\hb}Q\left(v_{i},\hb\mid\vb_{\neg C}\right)\log\frac{Q\left(v_{i},\hb\mid\vb_{\neg C}\right)}{P\left(v_{i},\hb\mid\vb_{\neg C}\right)}
\]
with respect to $Q\left(\v_{i}\mid\vb_{\neg C}\right)$ and $\left\{ Q\left(\h_{k}\mid\vb_{\neg C}\right)\right\} _{k=1}^{K}$.
This results in the following recursive relations:

\begin{eqnarray*}
Q\left(\v_{i}\mid\vb_{\neg C}\right) & \propto & \exp\left\{ G_{i}(\v_{i})+\sum_{k}H_{ik}(\v_{i})Q\left(\h_{k}\mid\vb_{\neg C}\right)\right\} ,\\
Q\left(\h_{k}\mid\vb_{\neg C}\right) & = & \frac{1}{1+\exp\{-w_{k}-\sum_{\v_{i}}H_{ik}(\v_{i})Q\left(\v_{i}\mid\vb_{\neg C}\right)-\sum_{j\in\neg C}H_{ik}(\v_{j})\}}.
\end{eqnarray*}

Now we make a further assumption that $\left|\sum_{\v_{i}}H_{ik}(\v_{i})Q\left(\v_{i}\mid\vb_{\neg C}\right)\right|\ll\left|\sum_{j\in\neg C}H_{ik}(\v_{j})\right|$,
e.g., when the set $\neg C$ is sufficiently large. This results in
$Q\left(\h_{k}\mid\vb_{\neg C}\right)\approx P\left(\h_{k}\mid\vb_{\neg C}\right)$
and
\[
Q\left(\v_{i}\mid\vb_{\neg C}\right)\propto\exp\left\{ G_{i}(\v_{i})+\sum_{k}H_{ik}(\v_{i})P\left(\h_{k}^{1}\mid\vb_{\neg C}\right)\right\} ,
\]
which is essentially the data model $P\left(\v_{i}\mid\hb\right)$
in Eq.~(\ref{eq:data-model}) with $h_{k}$ being replaced by $P\left(\h_{k}^{1}\mid\vb_{\neg C}\right)$.

The overall complexity of computing $Q\left(\v_{i}\mid\vb_{\neg C}\right)$
is the same as that of evaluating $P\left(\v_{i}\mid\vb_{\neg C}\right)$
in Eq.~(\ref{eq:predictive-distribution}). However, the approximation
is often numerically faster, and in the case of continuous variables,
it has the simpler functional form.